\title{Cardiac MR Image Segmentation Techniques: an overview}
\author{
        Tizita Nesibu Shewaye}
\date{\today}

\documentclass[letterpaper, 10 pt, conference]{ieeeconf}
\usepackage{graphicx}

\begin{document}
\maketitle

\begin{abstract}
Broadly speaking, the objective in cardiac image segmentation is to delineate the outer and inner walls of the heart to segment out either the entire or parts of the organ boundaries. This paper will focus on MR images as they are the most widely used in cardiac segmentation -- as a result of the accurate morphological information and better soft tissue contrast they provide. This cardiac segmentation information is very useful as it eases physical measurements that provides useful metrics for cardiac diagnosis such as infracted volumes, ventricular volumes, ejection fraction, myocardial mass, cardiac movement, and the like. But, this task is difficult due to the intensity and texture similarities amongst the different cardiac and background structures on top of some noisy artifacts present in MR images. Thus far, various researchers have proposed different techniques to solve some of the pressing issues. This seminar paper presents an overview of representative medical image segmentation techniques. The paper also highlights preferred approaches for segmentation of the four cardiac chambers: the left ventricle (LV), right ventricle (RV), left atrium (LA) and right atrium (RA), on short axis image planes.
\end{abstract}

\section{Introduction}
Cardiac Magnetic Resonance Imaging (MRI) provides accurate morphological information and better soft tissue contrast of a human heart. This image is very useful for early diagnosis of various heart abnormalities. In short, MRI has emerged as an excellent tool to analyze cardiac function, viability, and abnormality~\cite{Ringenberg2014}. In this regard, very useful information include infracted volumes, ventricular volumes, ejection fraction, myocardial mass, cardiac movement, and the like. This information entails accurate segmentation of the boundaries of the heart. Hence, cardiac segmentation is vital for diagnosis as well intervention planning as it provides accurate visualization of the subjects’ cardiac chambers. 

\begin{figure}[!ht]
	\centering
		\includegraphics[width=0.5\textwidth]{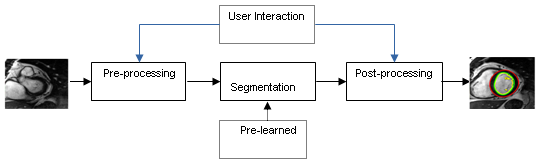}
	\caption{\textit{a generic cardiac MRI segmentation framework.}}
	\label{fig:img001}
\end{figure}

One way of obtaining this segmentation information is manually using trained experts. Unfortunately, this approach is time consuming, tiresome, and it is prone to inter- and intra-observer variability. During the past two decades, tremendous progress has been made to shift toward automated segmentation techniques based on automatic image processing techniques~\cite{Ringenberg2014}. Figure~\ref{fig:img001} shows a very generic cardiac MRI segmentation framework.

For a given MRI slice, the framework begins with a pre-processing step. The pre-processing step could encompass simple noise filtering (smoothing) step, image region of interest (ROI) selection (either automatically or with the help of a user interaction). The next step is the actual segmentation step. Since we are considering automatic segmentation cases here within, there is no user interaction in this step but, the segmentation could rely on models learned a-priori. Then follows a post-processing step. In the post-processing step, errors are corrected and unnecessary regions are discarded. This step could also involve user interaction for correcting these errors. This last step is often referred as a refinement step. 

This paper provides an overview of image segmentation techniques prevalently used in the literature for cardiac MRI segmentation. The presentation will be constrained to short axis MRI conforming to the literature. We also provide an overview of cardiac MRI segmentation techniques in the literature. The objective is to provide material that could be used a starting point for beginning researchers who aspire to know about cardiac MRI segmentation coarsely.

\section{The Human Heart and Cardiac MRI}
\begin{figure}
	\centering
		\includegraphics[width=0.45\textwidth]{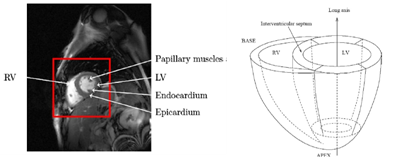}
	\caption{\textit{short-axis cardiac MR image with ROI in red identifying the heart (left) and ventricle geometry (right)~\cite{Petitjean2011}.}}
	\label{fig:img002}
\end{figure}

The human heart contains four chambers, right and left atrium, which are the upper chambers of the heart and right and left ventricles, which are the lower chambers of the heart. When compared to ventricles the atria are smaller in size, and have less muscular and thinner walls. Their function is to bring blood to the heart through the veins which are connected to it. In contrast ventricles are larger in size and stronger pumping chambers of the heart, which direct blood out of the heart through the arteries that are connected to it. The right side of the heart keeps pulmonary circulation to the nearby lungs whereas the left side of the heart maintains the systemic circulation by pumping blood to the whole body. Due to this the left side chambers are larger and have more myocardium in there wall than the right one. Figure~\ref{fig:img002} shows a schematic view of the heart ventricles slices along a short axis side by side with an actual MRI showing the different parts.

The human heart is constantly in motion. This motion necessitates a special acquisition sequence to capture the heart structures at the right moment. This heart motion problem is solved with the use of ECG gated imaging which allows for stop motion imaging by acquiring data only during a specified portion of the cardiac cycle. This is referred as cine MRI. In cardiac MRI the use of contrast agent in the blood stream helps segmentation of heart structures in what is called MR angiography (MRA). In prospective connotations, cardiac MR imaging are acquired using either cine MRI, MRA, or a combination of both techniques.

\section{Image Segmentation Techniques}
Image segmentation refers to the process of partitioning a digital image into multiple segments i.e. set of pixels, pixels in a region are similar according to some homogeneity criteria such as color, intensity or texture, so as to locate and identify objects and boundaries in an image~\cite{gonzalez2006}. In a more formal way, if X represents the entire spatial region occupied by an image, then image segmentation is a process of partitioning  X into n sub regions, $X_1$, $X_2, \cdots, X_n$, such that: (a) $\bigcup^n_
{i=1}$ $X_i=X$; (b) $X_i$ is a connected set, $i=1,2,\cdots,n$; (c) $X_i\cap X_j$= $\phi$ for all i and j,$i\neq j$; (d)$ P (X_i) = True$ for $i= 1,2,\cdots,n$; (e) $P (X_i\cup X_j) = False$ for any regions $X_i$ and $X_j$; $\phi$ denotes the empty set, and $P (X_i)$ is a logical predicate that evaluates to true for  each distinct region based on the homogeneity evaluation of the region. Condition (e) states that if two different regions were merged then the homogeneity of the resulting region would seize to exist, meaning in the segmented image, regions must be both homogeneous and maximal. 

There is plethora of segmentation techniques proposed in the literature throughout the years by various researchers all over the world. But, ironically, there is not a single method which outperforms the others under different images, in fact, all methods are not equally good for different images~\cite{Singh2010}. In this section, we recap the most basic segmentation techniques in a generic manner.

\subsection{Region based segmentation}
Region growing: The idea is that first a seed point is chosen and then all adjacent pixels are merged to this pixel if they fulfill the predetermined uniformity criteria, justifying the term growing. For example, in an RGB image, one can use the Euclidean distance in RGB space as a criterion between adjacent pixels. If pixel at $(i, j)$ has color $C (i, j)$and pixel at $(i+1, j+1)$ has color $C'(i+1, j+1)$ then the Euclidean distance between the two points can be represented as follows:

\begin{equation}
\centering
D'{}^2 (i,j)=(R_c-R_c')^2+ (G_c- G_c' )^2+ (B_c- B_c' )^2
\end{equation}
Region growing is similar with Split and Merge (discussed momentarily) in the limiting case when the splitting is done down to a single pixel. Every time a pixel is added to the region the mean values of the region are updated taking into consideration the newly added point. The region growing algorithm can be accomplished with the following series of steps:
(i) start with a seed pixel; (ii) using the 8 - connectivity model, calculate the uniformity measure of each neighboring pixel with the seed point; (iii) if the uniformity measure is satisfied, add the pixel to this region; (iv) label the added pixel according to the selected labeling color for this region and update the statistics of the region taking into account the newly added member; (v) when growing for this region stops, as denoted by a false uniformity measure for all neighboring regions, pick the next available unlabeled pixel as a seed point and continue with the above series of steps. Since the region growth starts growing from single points, it is common to get regions with only one pixel. In the final image these pixels will stand out clearly as they are labeled with distinct colors and resemble ‘salt and pepper’ noise. As post processing for this algorithm, the mask for the segmented image is filtered by a median filter. This will ensure that the one pixel regions will be merged into regions with most elements in their 8 - connectivity neighbors. 

\begin{figure}
	\centering
		\includegraphics[width=0.45\textwidth]{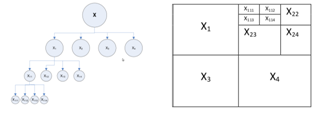}
	\caption{\textit{splitting procedure of split and merge segmentation as shown on image plane quad tree structure.}}
	\label{fig:img003}
\end{figure}

\textit{Split and Merge technique}: This technique can be seen as a generalization of region growing which consists of growing regions with large areas. In split and merge, the image is initially split into small regions satisfying the conditions of segmentation stated in section 1. Then these regions are merged into larger regions satisfying the segmentation conditions.\\
If X represents the entire image region, then the split and merge procedure is as follows: (i)Split into four quadrants any region $X_i$ for which $P_1 (X_i) = False$; (ii) When no further splitting is possible start merging adjacent regions $X_i$ and $X_j$ for which $P_2  (X_i\cup X_j) = True$; (iii) Stop when no further merging is possible; (iv) Label each region with a distinct color to differentiate each region and aid visual perception. Note here that P1 and P2 could be the same or different predicates that take into account different statistical criterion. The splitting procedure of this method can be represented in a quad tree structure as shown in Figure~\ref{fig:img003}.

\subsection{Clustering techniques}
Clustering-based approaches attempt to group together pixel points based on a similarity measure without considering spatial relationship (neighborhood occurrences). The simplest case is, for example, binarizing a given image, which is basically segmentation into two classes, using a single threshold. In this case, all pixels with gray scale value less or equal to the threshold would be assigned to one class, and the rest would be assigned to the second class. In the literature, there are many variants of these clustering techniques that employ either a single or a range of threshold values, determined in various ways, to segment images into distinct classes. The most widely used approaches are highlighted here below.

\subsubsection{\textit{Histogram based thresholding}} In this approach, a histogram is constructed on the feature space one would like to perform the segmentation. The feature space could be as simple as gray scale values or perhaps a multidimensional texture descriptor feature. Then, threshold values are automatically set to the valleys occurring along the histogram. This technique would yield good clustering of similar pixels as long as the histogram is well defined with peaks and valleys. Otherwise, it would lead to over-segmentation in the case of multiple valley occurrences. \\

\subsubsection{\textit{Gaussian Mixture Models (GMM)}} 
\begin{figure}
	\centering
		\includegraphics[width=0.45\textwidth]{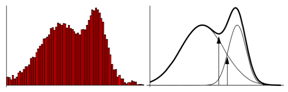}
	\caption{\textit{Image histogram and its approximation (thick curve) using two Gaussian distributions (thin curves)~\cite{Maintz2002}.}}
	\label{fig:img004}
\end{figure}

Once we have obtained histogram of the pixel values in a feature space, rather than setting threshold values based on valley positions, we can model the observed distribution using a mixture of Gaussian distributions. A GMM is a weighted sum of K component Gaussian densities expressed as: $p(x|\lambda) = \sum^K_{i=1} w_i g(x|\mu_i,\sum_i)$, where x is a D-dimensional continuous-valued data vector (i.e. measurement or features),$ w_{i,i} = 1, \cdots, K$, are the mixture weights, and $g(x|\mu_i,\sum_i)$, are the component Gaussian densities. Figure~\ref{fig:img004} depicts approximation of a given histogram with a mixture of two Gaussian distributions.

Given a histogram, the parameters of the mixture Gaussian distributions can be estimated based on maximum likelihood estimation which is the most popular and well established method for this kind of problems~\cite{Reynolds2009}. Once the GMM is determined, the segmentation is carried out by assigning the label (l) that maximizes the conditional probability $p(x=l│M)$ for each pixel. A shortcoming of this approach is that the number of mixture component, K, needs to be specified a-priori or another algorithm capable of determining sufficient model complexity (provide regularization) need to be used.

\subsubsection{\textit{K-means clustering}~\cite{macqueen1967}} K-means clustering based segmentation is an iterative algorithm that divides image pixels into K distinct classes using K-1 threshold values. The threshold values are determined in such a way to minimize the total with-in class variance. This algorithm can be summarized as follows: (i) initialize K-1 thresholds arbitrarily; (ii) segment the image according to the set thresholds; (iii) for each class/segment, compute the 'cluster center'; (iv) for each segment, compute the mean pixel value (in the feature space); (v) reset the cluster centers to the computed mean values; (vi) reset the thresholds to be midway between the cluster centers, and segment the image; (vii) iterate until the cluster centers do not move anymore. Again similar to the GMM approach, its main downside is that K must be known or set a priori. 

\subsection{Edge based segmentation}
Another approach for image segmentation is the edge based segmentation which tries to separate regions within the image by finding their edges. In its simplest form, it starts out with an edge detection algorithm, for example, using Sobel operator, canny edge detector, or the like. Then it processes the edge image so that only closed object boundaries remain. Finally, the results are transformed into an ordinary segmented image by filling in the object boundaries. This technique is very sensitive to noise in the image. For noisy image, the edge detection step would lead to many spurious and discontinuous edges. This can be partially mediated by utilizing an efficient edge linking technique to obtain closed boundaries that conform to the underlying image regions. In the literature, successful edge linking approaches are based on the Hough transform, neighborhood search, and network analysis. The watershed algorithm is one such segmentation technique based on this edge notion and watershed transformation ~\cite{Maintz2002}. 

\subsection{Trained classifier based techniques}
In this approach, a classifier is trained beforehand using annotated training samples. Then, the classifier is made to label a test image (an unseen instance) into distinct segments. A popular classifier for segmentation in the literature is Neural Network. Neural networks are massively connected networks of elementary processors that can learn non linear relationships between input and outputs. For segmentation, neural networks can be trained directly on the input image pixel feature values~\cite{Blanz1990}, or can also be trained on input image histograms to output relevant threshold values~\cite{Babaguchi1990}.

\subsection{Active contour technique}
An active contour, also known as snake, is  a curve defined in an image that is allowed to change its location and shape until it best satisfies predefined conditions~\cite{Maintz2002}. It is usually used to segment an object by letting it settle around the object boundary. Maintz~\cite{Maintz2002} models a snake $C$ as a parameterized curve $C(s)=(x(s),y(s))$, where the parameter s varies from 0 to 1 in such a way $C(0)$ gives the coordinate of the starting point and $C(1)$ gives the end coordinate. The movement of the snake is modeled as an energy minimization process, where the total energy $E$ to be minimized consists of three terms:

\begin{equation}
\centering
E= \int_0^1 (E_i (C(s))+ E_e (C(s))+ E_c (C(s)))ds
\end{equation}

The term $E_i$ is based on internal forces of the snake; it increases if the snake is stretched or bent. The term $E_e$ is based on external forces; it decreases if the snake moves closer to a part of the image we wish it to move to. For example, if we wish the snake to move to edges, we may base this energy term on edgeness values. The last term $E_c$ can be used to impose additional constraints, such as penalizing the creation of loops in the snake, penalizing moving too far away from the initial position, or penalizing moving into an undesired image region. A modification of this technique which embeds the curve into a higher dimensional function such that at time zero the initial contour corresponds to the level zero is known as level set segmentation~\cite{Osher1988}. This method is sometimes known as implicit snake. Its advantages include it being implicit and parameter free.

\subsection{Atlas-based methods}
Atlas based segmentation involves constructing a set of registered and manually annotated segmentations of an object, which is called an Atlas, a-priori. Then for a given image instance, the segmentation begins first by registering the image in accordance with the set of images in the Atlas. Consequently, the segmentation label of each pixel is inferred from the corresponding labels, in that position, in the image instances of the Atlas.  Common ways to infer the label include majority voting scheme, weighted voting scheme~\cite{Isgum2009}, or probabilistic maximum a posteriori inference~\cite{Depa2010}. Atlas based segmentation has proved to be very useful and better when there is significant intra-class variation of the target region to be segmented~\cite{Depa2010}.

\section{Segmentation of Human Heart Chambers}
This section presents segmentation techniques applied to the different heart chambers along with their specific characteristics.
 
\subsection{Left Ventricle (LV)}
Automatic segmentation of the LV in cine MR is challenging due to~\cite{Lu2009}: variation of its shape across slices and phases; intensity overlap within the cardiac regions; lack of edge information; inter-subject variability. Different authors have proposed different methods to overcome these challenges and result in accurate segmentation of LV and surrounding wall structures.

Lu et al.~\cite{Lu2009} proposed an automatic LV segmentation method using rudimentary image processing techniques. They begin by binarizing the given image ROI containing the heart structures using Otsu's threshold. Then all objects smaller than a predefined number of pixels are removed and the convex hull is computed for the surviving objects. Of all the remaining objects, the LV is regarded as the one with the largest roundness. Additionally, the LV blood pool contour, endocardial contour, papillary muscles and trabeculations' contours, and epicardial contours are detected sequentially through a series of masking, and morphological operations. This method though seems to work on the demonstrated sequences, makes strong geometrical assumptions on the structure of the segmented parts. Also it assumes a bimodal gray scale histogram as it uses Otsu's method which could be violated easily in presence of illumination variations and/or noise.

The majority of works on LV segmentation utilize active contour~\cite{Mitchell2001,Zhang2010} and atlas based segmentation~\cite{Zhuang2008, Lorenzo2004}. These approach, though have time consuming prior model learning, lead to exceptional results except whenever there is substantial variation in test cases. To utilize these techniques, the obtained results would have to of course offset the time-consuming prior model preparation and slow segmentation run time. On other cases, other authors have proposed LV segmentation techniques based on supervised Artificial Neural Network, clustering techniques using GMM and K-means clustering (for an in-depth review please refer to~\cite{Petitjean2011}. Again the most promising results are obtained using active counters and atlas based approaches~\cite{Petitjean2011}.
\begin{figure*}
	\centering
		\includegraphics[width=0.85\textwidth]{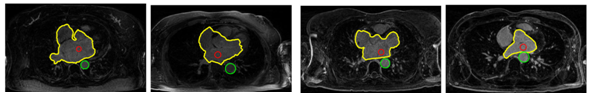}
	\caption{\textit{Illustration of Left Atrium segmentation~\cite{Zhu2012}. The TA is marked with green circle and the seed region with red circle. The yellow contour encloses the heart region.}}
	\label{fig:img005}
\end{figure*}

\subsection{Right Ventricle (RV)}
The segmentation problem is especially challenging when referring to the right ventricle (RV). due to the highly variable and crescent structure, thin and often indistinguishable myocardial walls, and inhomogeneous and ill-defined boundaries~\cite{Ringenberg2014}. RV segmentation algorithms are rare and have never been evaluated on common benchmark data~\cite{Ringenberg2014}. Due to its nature, the most successful approaches are based on multi-atlas segmentation framework~\cite{Zuluaga2012,Ou2012}.  For example, in Zuluaga et al.~\cite{Zuluaga2012} a target image is registered to all the intensity images of the atlases, after which the labeled images from the atlases are mapped back onto the target image. Then, all of the transformed labeled images are combined into a single segmentation through a label fusion method. Recently, ~\cite{Ringenberg2014} have reported marked improvements in speed and accuracy using a combination of a novel window-constrained accumulator thresholding technique, binary difference of Gaussian (DoG) filters, optimal thresholding, and morphology to drive the segmentation.

\subsection{Left Atrium (LA)}
The simplest approach based on cardiac MRA focus on segmentation of the left atrium by first extracting the whole blood pool by intensity thresholding and then separating it into different heart chambers by making cuts at narrowing~\cite{John2005}. But, this method has the downside that it requires several thresholds to be set manually to be applicable for varying intensity distributions across patients. Also bear in mind that, if the narrowing between the chambers are not distinct, the approach is bound to fail. In addition, the authors have shown this method works well when coupled with user interaction to provide a seed point within the LA and then using region growing to segment the chamber. Depa et al.~\cite{Depa2010} segment the left atrium using robust atlas based segmentation. In their work, the authors use 15 manually segmented and registered templates as an Atlas. The actual segmentation is then carried out via a label fusion algorithm based on maximum a posterior estimate for each voxel of the test images. The results reported outperform other Atlas based variants.

Zhu et al.~\cite{Zhu2012} tackle this problem in two steps. First, they focus on automatically detecting the thoracic aorta since it can be easily detected using Hough Transform technique thanks to its consistent circular shape (in axial view). Given the location of the thoracic aorta, the second step initializes a seed region in the left atrium using region and gradient information and proceeds with an active contour initialized in this region and driven by gradient based information. Qualitative results are shown in Figure~\ref{fig:img005}.

\subsection{Right Atrium (RA)}
The works in the literature for segmenting the right atrium basically rely on segmenting the blood pool in cardiac MRA similar to the left atrium segmentation discussed above. A notable exception is the works of Chen et al.~\cite{Chen2011}, whom use advanced level set based image segmentation to extract the right atrium. In their approach, rather than driving the level set iteration via image intensity and gradient solely, they incorporate a shape model constructed a-priori to guide the segmentation. The shape model is constructed using eigenshapes determined via Principal Component Analysis (PCA) from manually annotated and aligned shapes. During the level set optimization iteration, the energy minimization is also done with respect to each eigenshape's weights. Their results demonstrated notable improvement over variants that do not incorporate shape model.

\section{Segmentation Accuracy}
\begin{figure}
	\centering
		\includegraphics[width=0.45\textwidth]{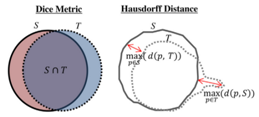}
	\caption{\textit{Visual illustration of segmentation evaluation metrics~\cite{Ringenberg2014}.}}
	\label{fig:img006}
\end{figure}

Segmentation accuracy measurement metrics is crucial to establish correctness of a segmentation algorithm. In cardiac MRI segmentation, two metrics are predominantly used: the Dice metric and the Hausdorff Distance. Dice metric computes the spatial overlap between two discretely labeled objects. On the other hand, the Hausdorff distance provides a maximal discrepancy between two labeled contours. Both are visually illustrated in Figure~\ref{fig:img006}. 

\section{Conclusive Summary}
To summarize, this paper briefly presented an overview of cardiac MRI segmentation techniques. Image segmentation has primarily two objectives. The first is to decompose an image into parts for further analysis and the second is to perform a change of representation. In short the objective is to simplify and or change the representation of an image into something that is more meaningful and easier to analyze. The choice of a particular segmentation technique to adopt depends on the nature of the image and problem considered. The four chambers of a human heart exhibit different characteristics when imaged. For example the left Atrium  is relatively small as compared to the left ventricle or neighboring anatomical structures; boundaries are not well defined when blood pool of the left atrium goes into the pulmonary veins; its shape varies significantly amongst different subjects making segmentation tasks very difficult.
Similarly, for the RA, its low contrast to noise ratio makes the boundaries between the RA and the nearby structures nearly indistinguishable. Since the shape of the cardiac ventricles is relatively simple and has low inter-patient variability, left ventricle and right ventricle segmentation has been thoroughly investigated in the literature.

Despite the many efforts by various researchers, there is still a lot of room for improvement especially with respect to LA/RA segmentations. It is worth emphasizing that looking at the literature, generally active counter methods and atlas-based segmentation techniques are at the front line. Combining different segmentation techniques to realize a coarse-to-fine segmentation hierarchy is also very promising~\cite{Ringenberg2014}.  For an in depth review, interested readers should refer to~\cite{Lu2009,Ringenberg2014}.


\bibliographystyle{plain}
\bibliography{referencepaper}
\end{document}